\documentclass[10pt,twocolumn,letterpaper]{article}
\usepackage[table]{xcolor}
\usepackage{iccv}
\usepackage{times}
\usepackage{epsfig}
\usepackage{amsmath}
\usepackage{amssymb}
\usepackage{booktabs}
\usepackage{graphicx}
\usepackage{float}
\usepackage{amssymb}
\usepackage{bbding}
\usepackage{pifont}
\usepackage{wasysym}
\usepackage{utfsym}
\usepackage{fontawesome}
\usepackage{color}
\usepackage{algorithm}
\usepackage{algorithmicx}  
\usepackage{algpseudocode}
\usepackage{comment}
\usepackage{tablefootnote}

\usepackage[pagebackref=true,breaklinks=true,letterpaper=true,colorlinks,bookmarks=false]{hyperref}
\def\mathbi#1{\textbf{\em #1}}

\iccvfinalcopy 

\def\iccvPaperID{3746} 

\ificcvfinal\fi
\begin{document}

\title{Exploring  Object-Centric Temporal Modeling for Efficient Multi-View 3D Object Detection}
\author{
Shihao Wang$^{1\dagger}$ \hspace{0.35cm} 
Yingfei Liu$^2$ \hspace{0.25cm} Tiancai Wang$^2$ \hspace{0.25cm} Ying Li$^1$ \hspace{0.25cm} 
Xiangyu Zhang$^2$ \\ [.5ex] $^1$Beijing Institute of Technology \hspace{0.9cm} $^2$MEGVII Technology \\ [.5ex]
}
\twocolumn[
{
\renewcommand\twocolumn[1][]{#1}%
\maketitle
\vspace{-0.25cm}
\begin{figure}[H]
\hsize=\textwidth
    \vspace{-1cm}
    \includegraphics[scale=0.525]{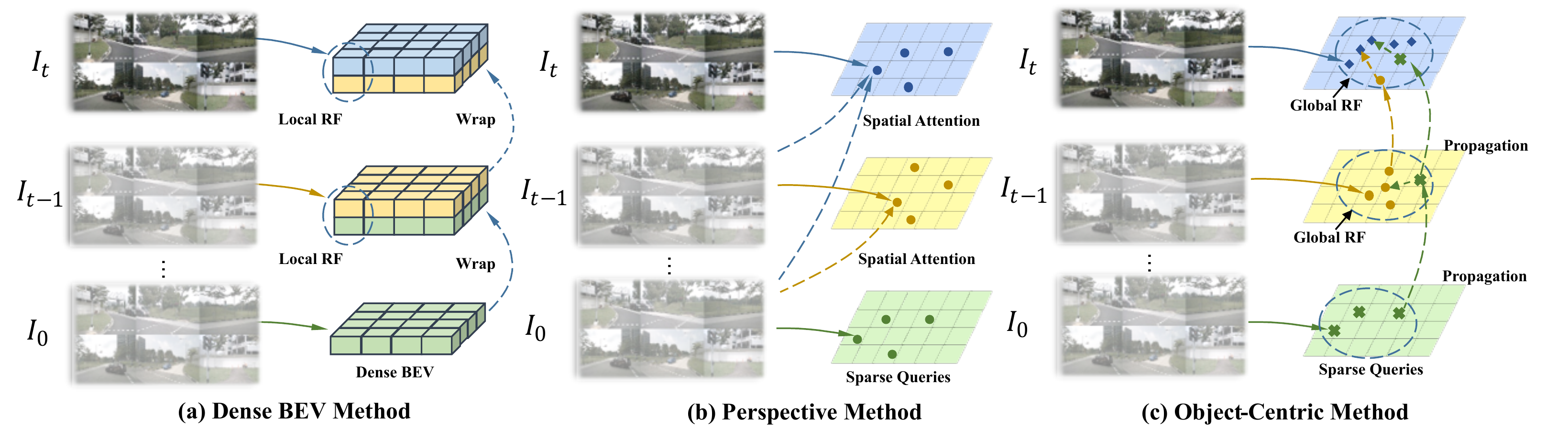}
    \caption{Different temporal fusion methods from bird-eye-view (BEV) space, perspective view, and our proposed object-centric. RF indicates receptive field. The solid lines and dotted lines represent spatial and temporal operations respectively.}
    \label{intro}
\end{figure}
}
]
\let\thefootnote\relax\footnotetext{$\dagger$ Work done during the internship at MEGVII Technology.}

\begin{abstract}
\vspace{-0.25cm}
In this paper, we propose a long-sequence modeling framework, named StreamPETR, for multi-view 3D object detection. Built upon the sparse query design in the PETR series, we systematically develop an object-centric temporal mechanism. The model is performed in an online manner and the long-term historical information is propagated through object queries frame by frame. Besides, we introduce a motion-aware layer normalization to model the movement of the objects. StreamPETR achieves significant performance improvements only with negligible computation cost, compared to the single-frame baseline. 
On the standard nuScenes benchmark, it is the first online multi-view method that achieves comparable performance  (67.6\% NDS \& 65.3\% AMOTA) with lidar-based methods. 
The lightweight version realizes 45.0\% mAP and 31.7 FPS, outperforming the state-of-the-art method (SOLOFusion) by 2.3\% mAP and 1.8$\times$ faster FPS. Code has been available at \href{https://github.com/exiawsh/StreamPETR}{\color{red} https://github.com/exiawsh/StreamPETR.git}.

\vspace{-0.5cm}
\end{abstract}

\section{Introduction}

Camera-only 3D detection is crucial for autonomous driving because of the low deployment cost and ease of detecting road elements. Recently, multi-view object detection has made remarkable progress by leveraging temporal information~\cite{li2022bevformer,huang2022bevdet4d,liu2022petrv2,li2022bevdepth,park2022time,lin2022sparse4d}. The historical features facilitate the detection of occlusion objects and greatly improve the performance. According to the differences between temporal representations, existing methods can be roughly divided into \emph{BEV temporal} and \emph{perspective temporal} methods.

BEV temporal methods~\cite{li2022bevformer,huang2022bevdet4d,li2022bevdepth,park2022time} explicitly warp BEV features from historical to current frame, as illustrated in Fig.~\ref{intro} (a),
where BEV features serve as an efficient intermediate representation for temporal modeling.
However, the highly structured BEV features limit the modeling of moving objects. This paradigm requires a large receptive field to alleviate this problem~\cite{huang2022bevdet4d,park2022time,li2022bevformer}. 

Different from these approaches,
perspective temporal methods ~\cite{liu2022petrv2,lin2022sparse4d} are mainly based on DETR~\cite{carion2020detr,zhu2020deformable}.  The sparse query design facilitates the modeling of moving objects ~\cite{lin2022sparse4d}. However, the sparse object queries need to interact with multi-frame image features for long-term temporal dependence (see Fig.~\ref{intro} (b)), leading to multiple computations.
Thus, existing works are either stuck in solving the moving objects or introducing multiple computation costs.

Based on the above analysis, we suppose it is possible to employ sparse queries as the hidden states of temporal propagation. In this way, we can utilize object queries to model moving objects while keeping high efficiency. 
Therefore, we introduce a new paradigm: \emph{object-centric temporal} modeling and design an efficient framework, termed StreamPETR, as shown in Fig.~\ref{intro} (c). StreamPETR directly performs frame-by-frame 3D predictions on streaming video. It is effective for motion modeling and is able to build long-term spatial-temporal interaction.

Specifically, a memory queue is first built to store the historical object queries. Then a propagation transformer conducts long-range temporal and spatial interaction with current object queries. The updated object queries are used to generate 3D bounding boxes and pushed into the memory queue. Besides, a motion-aware layer normalization (MLN) is introduced to implicitly encode the motion of the ego vehicle and surrounding objects at different time stamps. 

\begin{figure}[t]
\centering
\includegraphics[scale=0.09]{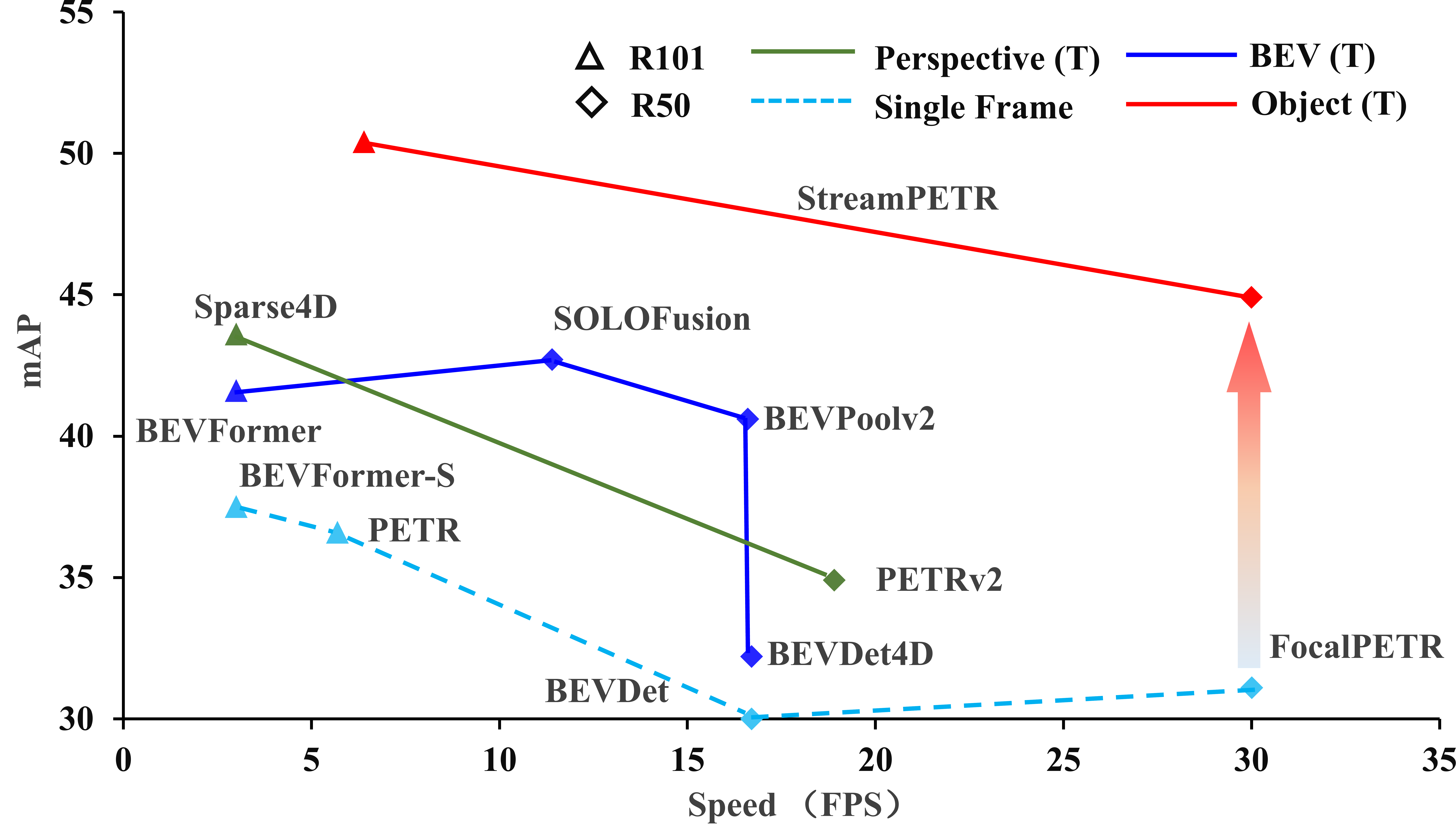}
\caption{The speed-accuracy trade-off of different models on nuScenes val set. The inference speed is calculated on RTX3090 GPU in online streaming video. T indicates the model with temporal modeling.}
\vspace{-0.5cm} 
\label{speed_vs_acc}
\end{figure}

Compared with existing temporal methods, the proposed object-centric temporal modeling brings several advantages. StreamPETR only processes a small number of object queries instead of dense feature maps at each time stamp, consuming negligible computational burden (as shown in Fig.~\ref{speed_vs_acc}). For moving objects, MLN alleviates the cumulative error in video streaming. Except for the location prior used in previous methods, StreamPETR additionally considers the semantic similarity by global attention, which facilitates the detection in motion scenes. To summarize, our contributions are:

\begin{itemize}
\item We pull out the key of streaming multi-view 3D detection and systematically  design an \emph{object-centric} temporal modeling paradigm. The long-term historical information is propagated through object queries frame by frame.

\item We develop an object-centric temporal modeling framework, termed StreamPETR. 
It models moving objects and long-term spatial-temporal interaction simultaneously, consuming negligible storage and computation costs.

\item  On the standard nuScenes dataset, StreamPETR outperforms all online camera-only algorithms. Extensive experiment shows that it can be well generalized to other sparse query-based methods, e.g. DETR3D~\cite{wang2022detr3d}. 
\end{itemize}

\section{Related Work}
\begin{figure*}[t]
	\centering  
     \includegraphics[width=0.95\linewidth]{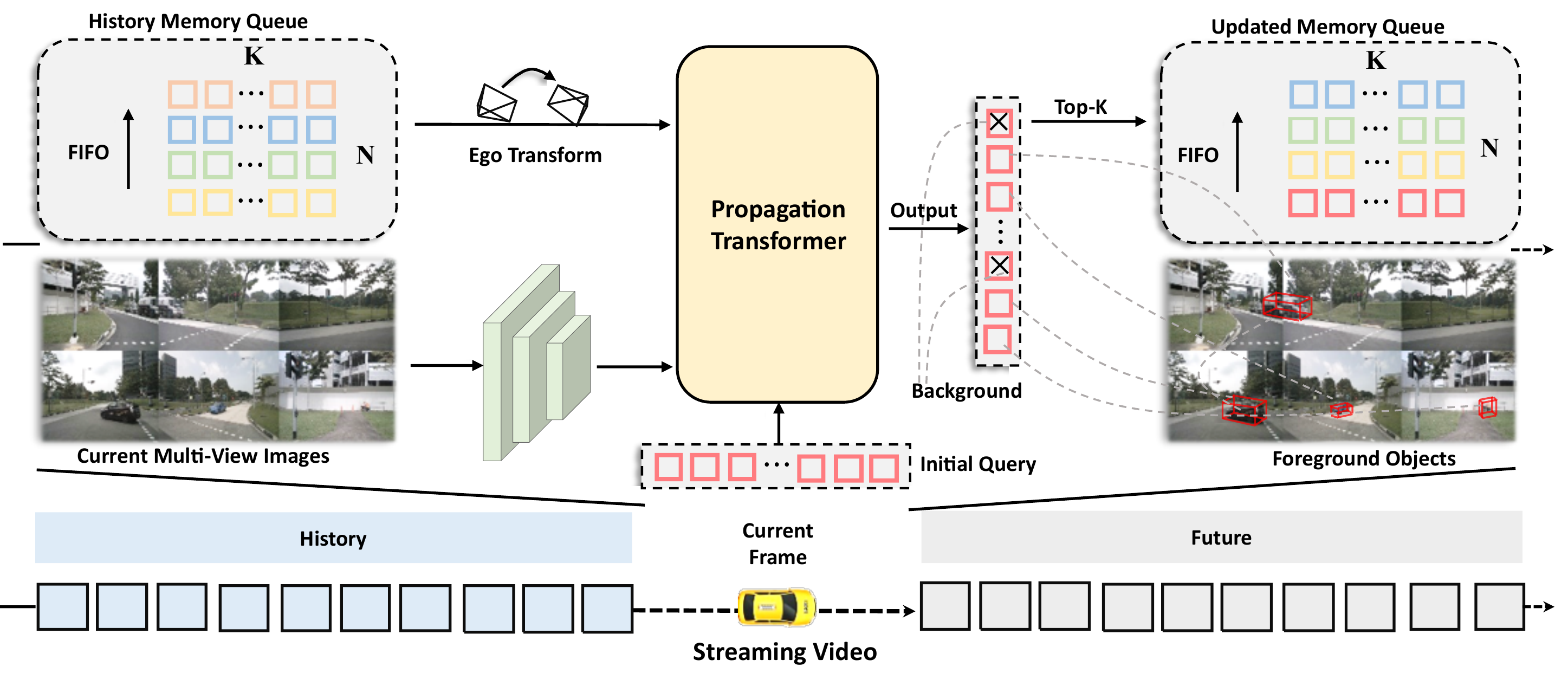}
	\caption{Overall architecture of the proposed StreamPETR. The memory queue stores the historical object queries. In the propagation transformer, recent object queries successively interact with historical queries and current image features to obtain temporal and spatial information. The output queries are further used to generate detection results and the top-K foreground queries are pushed into the memory queue. Through the recurrent update of the memory queue, the long-term temporal information is propagated frame by frame.}
	\label{architecture}
\vspace{-0.5cm} 
\end{figure*}

\subsection{Multi-view 3D Object Detection}
Multi-view 3D detection is an important task in autonomous driving, which needs to continuously process multi-camera images and predict 3D bounding boxes over time.
Pioneer's works~\cite{wang2022detr3d,liu2022petr,huang2021bevdet,li2022bevformer, jiang2022polarformer, wang2023object} focus on the efficient transformation from multiple perspective views to a unified 3D space in a single frame. The transformation can be divided into BEV-based methods~\cite{huang2021bevdet,li2022bevformer,xie2022m,huang2023fast,li2022bevdepth,jiang2022polarformer} and sparse query based methods~\cite{wang2022detr3d,liu2022petr,lin2022sparse4d,chen2022polar,wang2023object}. To alleviate the occlusion problem and ease the difficulty of speed prediction, recent works additionally introduce temporal information to extend these two paradigms. 

It is relatively intuitive to extend the single-frame BEV methods for temporal modeling.
BEVFormer~\cite{li2022bevformer} first introduces sequential temporal modeling into multi-view 3D object detection and applies temporal self-attention. BEVDet series~\cite{huang2022bevdet4d,li2022bevdepth, li2022bevstereo} use concatenate operation to fuse the adjacent BEV features and achieve remarkable results. Furthermore, SOLOFusion~\cite{park2022time} extends BEVStereo~\cite{li2022bevstereo} to long-term memory and reaches a promising performance. Without an intermediate feature representation, the temporal modeling of query-based methods is more challenging. PETRv2~\cite{liu2022petrv2} performs the global cross-attention, while DETR4D ~\cite{luo2022detr4d} and  Sparse4D~\cite{lin2022sparse4d} apply sparse attention to model the interaction between multi-frames, which introduce multiple computations. However, the sparse query design is convenient to model the moving objects~\cite{lin2022sparse4d}. In order to combine the advantages of the two paradigms, we utilize sparse object queries as the intermediate representation, which can model moving objects and efficiently propagate long-term temporal information.

\subsection{Query Propagation}
Since DETR~\cite{carion2020detr} is proposed in 2D object detection, the object query has been applied in many downstream tasks~\cite{zeng2022motr,meinhardt2022trackformer,zhang2022motrv2, zhang2022mutr3d,he2022queryprop} to model the temporal interaction. For video object detection, LWDN~\cite{jiang2019video} adopts a brain-inspired memory mechanism to propagate and update the memory feature.
QueryProp~\cite{he2022queryprop} performs query interaction to reduce the computational cost on non-key frames. It achieves significant improvements and maintains high efficiency. 3D-MAN~\cite{yang20213d} has a similar idea and extends a single-frame Lidar detector to multi-frames, which effectively combines the features coming from different perspectives of a scene. In object tracking, MOTR~\cite{zeng2022motr} and TrackFormer~\cite{meinhardt2022trackformer} propose the track query to model the object association across frames. MeMOT ~\cite{cai2022memot} employs a memory bank to build long temporal dependence, which further boosts performance. MOTRv2~\cite{zhang2022motrv2} eases the conflict between the detection and association tasks by incorporating an extra detector. MUTR~\cite{zhang2022mutr3d} and PF-Track~\cite{pang2023standing} extend MOTR~\cite{zeng2022motr} into multi-view 3D object tracking and achieve a promising result.

\section{Delving into Temporal Modeling}\label{analysis}
To facilitate our study, we present a generalized formulation for various temporal modeling designs. Given the perspective view features $F_{2d} = \{F^{0}_{2d}\cdots  F^{t}_{2d}\}$, dense BEV features  $F_{bev} = \{F^{0}_{bev}\cdots F^{t}_{bev}\}$ and sparse object features $F_{obj} = \{F^{0}_{obj}\cdots F^{t}_{obj}\}$. The dominant temporal modeling methods can be formulated as:
\begin{equation}\label{eq4}
\begin{aligned}
\tilde{F}_{out}\!=\!\varphi(F_{2d}, F_{bev}, F_{obj})
\end{aligned}
\end{equation}
where $\varphi$ is the temporal fusion operation, $\tilde{F}_{out}$ is the output feature that includes temporal information. We first describe the existing temporal modeling from BEV and perspective view. After that, the proposed object-centric temporal modeling is elaborated.
\begin{figure*}[thb]
	\centering  
	\includegraphics[width=1.0\linewidth]{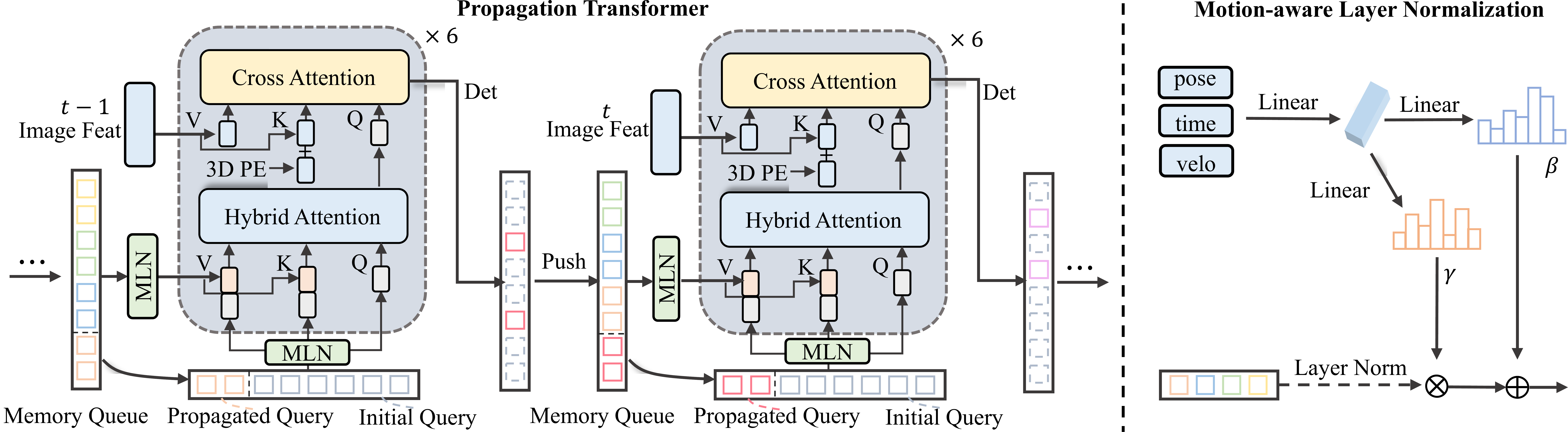}
	\caption{The details of the propagation transformer and motion-aware layer normalization. In the propagation Transformer~\cite{vaswani2017attention}, object queries interact with hybrid queries and image features iteratively. The motion-aware layer normalization encodes the motion attributes  (ego pose, timestamps, velocity) and performs a compensation implicitly. Rectangles of varying hues symbolize queries from distinct frames, gray rectangles represent initialized queries of current frame, dashed rectangles correspond to background queries.}
	\label{transformer}
\vspace{-0.5cm}
\end{figure*}

\noindent\textbf{BEV Temporal Modeling} uses the grid-structured BEV features to perform the temporal fusion. To compensate for the ego vehicle motion, the last frame feature $F^{t-1}_{bev}$ is usually aligned to the current frame. 
\begin{equation}\label{eq_bevdet4d}
\begin{aligned}
\tilde{F}^{t}_{bev} = \varphi(F^{t-1}_{bev}, F^{t}_{bev})
\end{aligned}
\end{equation}
Then a temporal fusion function $\varphi$  (concatenation~\cite{huang2022bevdet4d, li2022bevdepth} or deformable attention~\cite{li2022bevformer}) can be applied for intermediate temporal representation $\tilde{F}^{t}_{bev}$. Extending the above process to long temporal modeling, there are two main routes. The first one is to align the historical $k$ BEV features and concatenate them with the current frame.
\begin{equation}\label{eq_solofusion}
\begin{aligned}
\tilde{F}^{t}_{bev} = \varphi(F^{t-k}_{bev}, \cdots, F^{t-1}_{bev}, F^{t}_{bev})
\end{aligned}
\end{equation}
For another one, the long-term historical information is propagated through the hidden states of BEV features $\tilde{F}^{t-1}_{bev}$ in a recurrent manner. 
\begin{equation}\label{eq_bevformer}
\begin{aligned}
\tilde{F}^{t}_{bev} = \varphi(\tilde{F}^{t-1}_{bev}, F^{t}_{bev})
\end{aligned}
\end{equation}

However, the BEV temporal fusion only considers the static BEV features and ignores the movement of the objects, leading to spatial dislocation.

\noindent\textbf{Perspective Temporal Modeling} is mainly performed via interactions between object queries and perspective features. The temporal function $\varphi$ is usually achieved by the spatial cross-attention~\cite{liu2022petrv2, lin2022sparse4d, luo2022detr4d}:

\begin{equation}\label{eq5}
\begin{aligned}
\tilde{F}^{t}_{obj} = \varphi(F^{t-k}_{2d}, {F}^{t}_{obj}) \cdots + \varphi(F^{t}_{2d}, F^{t}_{obj})
\end{aligned}
\end{equation}

The cross-attention between object query and multi-frame perspective view requires repeated feature aggregation. Simply extending to long-term temporal modeling greatly increases the computation cost.

\noindent\textbf{Object-centric Temporal Modeling} is our proposed object-centric solution, which models the temporal interaction by object queries.
Through object queries, the motion compensation can be conveniently applied based on estimated states ${F}^{t-1}_{obj}$. 

\begin{equation}\label{eq_compensation}
\begin{aligned}
\tilde{F}^{t-1}_{obj} = \mu({F}^{t-1}_{obj} , M)
\end{aligned}
\end{equation}

where $\mu$ is an explicit linear velocity model or implicit function to encode motion attributes $M$ (including the relative time interval $\triangle t$, estimated velocity $v$, and ego-pose matrix $E$, which are the same definition in Sec.~\ref{sec_method}).
Further, a global attention $\varphi$ is constructed to propagate temporal information through object queries frame by frame:

\begin{equation}\label{eq_object}
\begin{aligned}
\tilde{F}^{t}_{obj} = \varphi(\tilde{F}^{t-1}_{obj}, F^{t}_{obj})
\end{aligned}
\end{equation}

\section{Method}
\label{sec_method}
\subsection{Overall Architecture}
As illustrated in Fig.~\ref{architecture}, StreamPETR is built upon end-to-end sparse query-based 3D object detectors~\cite{liu2022petr, wang2022detr3d}. It consists of an image encoder, a recursively updated memory queue, and a propagation transformer~\cite{vaswani2017attention}. The image encoder is a standard 2D backbone, which is applied to extract semantic features from multi-view images. Then the extracted features, information in the memory queue, and object queries are fed into the propagation transformer to perform the spatial-temporal interaction. The main difference between StreamPETR and single-frame baseline is the memory queue, which recursively updates the temporal information of object queries. Combined with the propagation transformer, the memory queue can propagate temporal priors from previous to current frames efficiently.

\subsection{Memory Queue}
We design a memory queue of $N\times K$ for effective temporal modeling. $N$ is the number of stored frames and $K$ is the number of objects stored per frame. According to the experience, we set $N=4$ and $K=256$ (ensuring high recall in complex scenarios). After the preset time interval $\tau$, the relative time interval $\triangle t$, context embedding $Q_{c}$, object center $Q_{p}$, velocity $v$, and ego-pose matrix $E$ of selected object queries are stored in memory queue. Specifically, the above information, corresponding to foreground objects (with top-$K$ highest classification score), is selected and pushed into the memory queue. The entrance and exit of the memory queue follow the first-in, first-out (FIFO) rule. When information from a new frame is added to the memory queue, the oldest is discarded. Actually, the proposed memory queue is highly flexible and customized, users can freely control the maximal memory size $N\times K$ and saving interval $\tau$ during both training and inference.

\subsection{Propagation Transformer}
\label{propagation_transformer}
As illustrated in Fig.~\ref{transformer}, the propagation transformer consists of three main components:  (1) the motion-aware layer normalization module implicitly updates the object state according to the context embedding and motion information recorded in the memory queue; (2) the hybrid attention replaces the default self-attention operation. It plays the role of temporal modeling and removing duplicated predictions; (3) the cross-attention is adopted for feature aggregation. It can be replaced with an arbitrary spatial operation to build the relationship between image tokens and 3D object queries, such as global attention in PETR~\cite{liu2022petr} or sparse projective attention in DETR3D~\cite{wang2022detr3d}.

\noindent\textbf{Motion-aware Layer Normalization}
is designed to model the movement of objects. For simplicity, we take the transformation process from the last frame $t-1$ as the example and adopt the same operation for other previous frames. Given the ego pose matrix from the last frame $E_{t-1}$ and current frame $E_{t}$, the ego transformation $E_{t-1}^{t}$ can be calculated as:
\begin{equation}
\label{explicit_align_1}
E_{t-1}^{t} =\ E_{t}^{inv}\cdot\ E_{t-1} 
\end{equation} 
Assume that objects are static, 3D centers $Q_{p}^{t-1}$ in memory queue can be explicitly aligned to the current frame, which is formulated as:

\begin{equation}
\label{explicit_align_2}
\tilde{Q}_{p}^{t} = E_{t-1}^{t} \cdot Q_{p}^{t-1}
\end{equation}

\begin{table*}
\centering
\caption{Comparison on the nuScenes \texttt{val} set. ${}^\ast$Benefited from the perspective-view pre-training. ${}^\ddagger$ 300 randomly initialized queries and 128 propagation queries. ${}^\dag$ Offline method using future frames. FPS is measured on RTX3090 with fp32.}
\label{tab:main_val_set}
\tiny
\resizebox{\textwidth}{!}{
\setlength{\tabcolsep}{3.5pt}
\begin{tabular}{l|c|c|c|c|c|c|c@{\hspace{1.0\tabcolsep}}c@{\hspace{1.0\tabcolsep}}c@{\hspace{1.0\tabcolsep}}c@{\hspace{1.0\tabcolsep}}|c} 

\toprule
\textbf{Methods} & \textbf{Backbone} & \textbf{Image Size}  & \textbf{Frames} & \textbf{mAP}$\uparrow$  &\textbf{NDS}$\uparrow$  & \textbf{mATE}$\downarrow$ & \textbf{mASE}$\downarrow$   &\textbf{mAOE}$\downarrow$   &\textbf{mAVE}$\downarrow$   &\textbf{mAAE}$\downarrow$ &\textbf{FPS}$\uparrow$ \\
\midrule
BEVDet \cite{huang2021bevdet} & ResNet50 & 256 $\times$ 704 & 1                                     & 0.298 & 0.379 & 0.725 & 0.279 & 0.589 & 0.860 & 0.245 & 16.7\\ 
BEVDet4D \cite{huang2022bevdet4d} & ResNet50 & 256 $\times$ 704 & 2                                   & 0.322 & 0.457 & 0.703 & 0.278 & 0.495 & 0.354 & 0.206 & 16.7\\ 
PETRv2 \cite{liu2022petrv2} & ResNet50 & 256 $\times$ 704 & 2                                   & 0.349 & 0.456 & 0.700 & 0.275 & 0.580 & 0.437 & 0.187 & 18.9\\ 
BEVDepth \cite{li2022bevdepth} & ResNet50 & 256 $\times$ 704 & 2                                   & 0.351 & 0.475 & 0.639 & 0.267 & 0.479 & 0.428 & 0.198 &15.7\\ 
BEVStereo \cite{li2022bevstereo} & ResNet50 & 256 $\times$ 704 & 2                                  & 0.372 & 0.500 & 0.598 & 0.270 & 0.438 & 0.367 & 0.190 & 12.2\\ 
BEVFormerv2 \cite{yang2022bevformer} ${}^\dag\ast$ & ResNet50 & -  & -                               &   0.423&0.529  &0.618  &0.273  &0.413  &0.333  &0.188  &  -\\ 
SOLOFusion \cite{park2022time} & ResNet50 & 256 $\times$ 704 & 16+1                                 & 0.427 & 0.534 & 0.567 & 0.274 & 0.511 & 0.252 & 0.181 & 11.4\\ %
BEVPoolv2 \cite{huang2022bevpoolv2} & ResNet50 & 256 $\times$ 704  & 8+1                               &   0.406&0.526  &0.572  &0.275  &0.463  &0.275  &0.188  &  16.6\\ 
\rowcolor[gray]{.9} 
StreamPETR & ResNet50 & 256 $\times$ 704  & 8                               & 0.432 & 0.540 & 0.581 & 0.272 & 0.413 & 0.295 &  0.195 & 27.1\\ 
\rowcolor[gray]{.9} 
StreamPETR${}^\ast\ddagger$ & ResNet50 & 256 $\times$ 704  & 8                                  &0.450   &0.550  &0.613  &0.267  &0.413  &0.265  &0.196  & \textbf{31.7}  \\
\midrule 
DETR3D \cite{wang2022detr3d}${}^\ast$ & ResNet101-DCN & 900 $\times$ 1600  & 1                  & 0.349 & 0.434 & 0.716 & 0.268 & 0.379 & 0.842 & 0.200 & 3.7\\ 
Focal-PETR \cite{wang2022focal} & ResNet101-DCN & 512 $\times$ 1408 & 1                                & 0.390 & 0.461 & 0.678 & 0.263 & 0.395 & 0.804 & 0.202 & 6.6\\ 
PETR \cite{liu2022petr}${}^\ast$ & ResNet101-DCN & 512 $\times$ 1408 & 1                                & 0.366 & 0.441 & 0.717 & 0.267 & 0.412 & 0.834 & 0.190 & 5.7 \\ 
BEVFormer \cite{li2022bevformer}${}^\ast$ & ResNet101-DCN & 900 $\times$ 1600  & 4                   & 0.416 & 0.517 & 0.673 & 0.274 & 0.372 & 0.394 & 0.198 & 3.0\\ 
PolarDETR \cite{chen2022polar}-T${}^\ast$ & ResNet101-DCN & 900 $\times$ 1600  & 2                 & 0.383 & 0.488 & 0.707 & 0.269 & 0.344 & 0.518 & 0.196 & 3.5 \\ 
Sparse4D \cite{lin2022sparse4d}${}^\ast$ & ResNet101-DCN & 900 $\times$ 1600 & 4                 & 0.436 & 0.541 & 0.633 & 0.279 & 0.363 & 0.317 & \textbf{0.177}& 4.3\\ 
BEVDepth & ResNet101 & 512 $\times$ 1408 & 2                            & 0.412 & 0.535 & 0.565 & 0.266 & 0.358 & 0.331 & 0.190 &- \\ 
SOLOFusion & ResNet101 & 512 $\times$ 1408 & 16+1                          & 0.483 & 0.582 & \textbf{0.503} & 0.264 & 0.381 & \textbf{0.246} & 0.207 & - \\ 
\rowcolor[gray]{.9} 
StreamPETR${}^\ast$ & ResNet101 & 512 $\times$ 1408 & 8                                 &\textbf{0.504}  &\textbf{0.592}  &0.569  &\textbf{0.262}  &\textbf{0.315}  &0.257  &0.199 & 6.4 \\ 
\bottomrule
\end{tabular}}
\vspace{-0.25cm}
\end{table*}

where $\tilde{Q}_{p}^{t}$ is the aligned centers. Motivated by the task-specific control in generative model~\cite{dumoulin2016learned, park2019gaugan, zhang2023adding}, we adopt a conditional layer normalization to model the movement of the objects. As shown in Fig.~\ref{transformer}, the default affine transformation in layer normalization (LN) is closed. The motion attributes ($E_{t-1}^{t}, v, \triangle t$)  are flattened and converted to affine vectors $\gamma$ and $\beta$ by two linear layers ($\xi_1$, $\xi_2$):
\begin{equation}
    \label{motion-ln-mlp}
    \begin{aligned}
    \gamma &= \xi_1(E_{t-1}^{t}, v, \triangle t) ,\\   
    \beta &= \xi_2(E_{t-1}^{t}, v, \triangle t)
    \end{aligned}
\end{equation}

Afterward, the affine transformation is performed to get the motion-aware context embedding $\tilde{Q}_{c}^{t}$ and motion-aware position encoding 
$\tilde{Q}_{pe}^{t}$. 

\begin{equation}
    \label{motion-ln}
    \begin{aligned}
    \tilde{Q}_{pe}^{t}&=\gamma \cdot LN(\psi(\tilde{Q}_{p}^{t})) + \beta ,\\
    \tilde{Q}_{c}^{t} &=\gamma \cdot LN(Q_{c}^{t}) + \beta
    \end{aligned}
\end{equation}

where $\psi$ is a multi-layer perceptron (MLP) that converted the 3D sampled points $\tilde{Q}_{p}^{t}$ into position encoding $\tilde{Q}_{pe}^{t}$. For the sake of unification, the MLN is also adopted into current object queries. The velocity $v$ and time interval $\triangle t$ of the current frame are zero-initialized.

\noindent\textbf{Hybrid Attention layer.} 
The self-attention in DETR~\cite{carion2020detr} contributes to duplicated prediction removal. We replace it with hybrid attention, which additionally introduces temporal interaction. As shown in Fig.~\ref{transformer}, all stored object queries in the memory queue are concatenated with current queries to obtain the hybrid queries. The hybrid queries are regard as the $key$ and $value$ in multi-head attention. Since the number of hybrid queries is small (about 2$k$, which is far less than image tokens in the cross-attention), the hybrid attention layer brings negligible computation cost.

Following PETR~\cite{liu2022petr}, the $query$ can be defined as a randomly initialized 3D anchor. To fully utilize the spatial and context priors in streaming video, some object queries in the memory queue are directly propagated into the current frame. In our implementation, queries from the last frame are concatenated with randomly initialized queries. For a fair comparison, the number of randomly initialized queries and propagated queries are set to 644 and 256 respectively.

\begin{table*}
\centering
\caption{Comparison on the nuScenes \texttt{test} set. TTA is test time augmentaion.}
\label{tab:main_test_set}
\tiny
\resizebox{\textwidth}{!}{
\setlength{\tabcolsep}{4pt}
\begin{tabular}{l|c|c|c|c|c|c|c@{\hspace{1.0\tabcolsep}}c@{\hspace{1.0\tabcolsep}}c@{\hspace{1.0\tabcolsep}}c@{\hspace{1.0\tabcolsep}}c} 

\toprule
\textbf{Methods} &\textbf{Modality} & \textbf{Backbone} & \textbf{Image / Voxel} & \textbf{TTA} & \textbf{mAP}$\uparrow$  &\textbf{NDS}$\uparrow$  & \textbf{mATE}$\downarrow$ & \textbf{mASE}$\downarrow$   &\textbf{mAOE}$\downarrow$   &\textbf{mAVE}$\downarrow$   &\textbf{mAAE}$\downarrow$  \\
\midrule
CenterPoint \cite{yin2021center}  & L    & -  & 0.075$\times$0.075$\times$0.2  & \ding{56} & 0.603 & 0.673 & \textbf{0.262} & \textbf{0.239} & 0.361 & 0.288 & 0.136\\
\midrule
FCOS3D \cite{wang2021fcos3d}  & C    & R101-DCN  & 900 $\times$ 1600  & \ding{52} & 0.358 & 0.428 & 0.690 & 0.249 & 0.452 & 1.434 & 0.124 \\
DETR3D \cite{wang2022detr3d}  & C     & V2-99 & 900 $\times$ 1600 & \ding{52} & 0.412 & 0.479 & 0.641 & 0.255 & 0.394 & 0.845 & 0.133 \\
MV2D \cite{wang2023object}  & C    & V2-99      & 640 $\times$ 1600 & \ding{56} & 0.463 & 0.514 & 0.542 & 0.247 & 0.403 & 0.857 & 0.127 \\
UVTR \cite{li2022unifying}   & C      & V2-99      & 900 $\times$ 1600 & \ding{56} & 0.472 & 0.551 & 0.577 & 0.253 & 0.391 & 0.508 & 0.123 \\
BEVFormer \cite{li2022bevformer}  & C    & V2-99      & 900 $\times$ 1600 & \ding{56} & 0.481 & 0.569 & 0.582 & 0.256 & 0.375 & 0.378 & 0.126 \\
PETRv2 \cite{liu2022petrv2}   & C      & V2-99  & 640 $\times$ 1600 & \ding{56} & 0.490 & 0.582 & 0.561 & 0.243 & 0.361 & 0.343 & 0.120 \\

PolarFormer  \cite{jiang2022polarformer} & C  & V2-99      & 900 $\times$ 1600 & \ding{56} & 0.493 & 0.572 & 0.556 & 0.256 & 0.364 & 0.439 & 0.127 \\
BEVStereo  \cite{li2022bevstereo} & C    & V2-99      & 640 $\times$ 1600  & \ding{56} & 0.525 & 0.610 & 0.431 & 0.246 & 0.358 & 0.357 & 0.138 \\
\rowcolor[gray]{.9} 
StreamPETR & C   & V2-99 & 640 $\times$ 1600 & \ding{56} &0.550 &0.636  &0.479  &\textbf{0.239}  &0.317  &0.241 &\textbf{0.119}\\
\midrule
BEVDet4D \cite{huang2022bevdet4d} & C  & Swin-B~\cite{liu2021swin}     & 900 $\times$ 1600 & \ding{52} & 0.451 & 0.569 & 0.511 & 0.241 & 0.386 & 0.301 & 0.121 \\
BEVDepth  \cite{li2022bevdepth}  & C    & ConvNeXt-B & 640 $\times$ 1600 & \ding{56} & 0.520 & 0.609 & 0.445 & 0.243 & 0.352 & 0.347 & 0.127 \\
AeDet  \cite{feng2022aedet}  & C    & ConvNeXt-B & 640 $\times$ 1600 & \ding{52} & 0.531 & 0.620 & 0.439 & 0.247 & 0.344 & 0.292 & 0.130 \\
PETRv2     & C    & RevCol-L~\cite{cai2022reversible}  & 640 $\times$ 1600 & \ding{56} & 0.512 & 0.592 & 0.547 & 0.242 & 0.360 & 0.367 & 0.126 \\
SOLOFusion  \cite{park2022time} & C   & ConvNeXt-B & 640 $\times$ 1600& \ding{56} & 0.540 & 0.619 & 0.453 & 0.257 & 0.376 & 0.276 & 0.148 \\
\rowcolor[gray]{.9} 
StreamPETR   & C   & ViT-L & 800 $\times$ 1600 & \ding{56} &\textbf{0.620} &\textbf{0.676} &0.470  &0.241  &\textbf{0.258}  &\textbf{0.236}  &0.134 \\
\bottomrule
\end{tabular}}
\vspace{-0.5cm}
\end{table*}

\section{Experiments}
\subsection{Dataset and Metrics}
We evaluate our approach on the large-scale NuScenes dataset~\cite{caesar2020nuscenes} and Waymo Open dataset~\cite{sun2020scalability}.

\noindent\textbf{The nuScenes Dataset} includes 1000 scenes, which are 20 seconds in length and annotated at 2Hz. The camera rig covers the full 360° field of view (FOV). The annotations contain up to 1.4M 3D bounding boxes, and 10 common classes are used for evaluation: car, truck, bus, trailer, construction vehicle, pedestrian, motorcycle, bicycle, barrier, and traffic cone. We compare the methods with the following metrics, the nuScenes Detection Score (NDS), mean Average Precision (mAP), and 5 kinds of True Positive (TP) metrics including average translation error (ATE), average scale error (ASE), average orientation error (AOE), average velocity error (AVE), average attribute error (AAE). Following the standard evaluation
metrics, we report the average multi-object tracking accuracy (AMOTA), average multi-object tracking precision
(AMOTP), recall (RECALL), multi-object tracking
accuracy (MOTA) and ID switch (IDS) for 3D object tracking task.

\noindent\textbf{Waymo Open Dataset} collects camera data only spanning a horizontal FOV of ~230 degrees. The ground truth bounding boxes are annotated to a maximum range of 75 meters. The longitudinal error tolerant metrics LET-3D-AP, LET-3D-AP-H and LET-3D-APL are used for evaluation. Noting that we only use 20\% of training data for fair comparison according to common practice.
\vspace{-0.25cm}
\begin{table}
\centering
\caption{Comparison of 3D object tracking on
nuScenes test set.} 
\label{tab:tracking}
\tiny
\resizebox{0.475\textwidth}{!}{
\setlength{\tabcolsep}{3pt}
\vspace{1em}
\begin{tabular}{c|c|c|c|c}


\toprule
\textbf{Methods} & 
\textbf{AMOTA$\uparrow$} & \textbf{AMOTP}$\downarrow$  &\textbf{RECALL}$\uparrow$ & \textbf{IDS}$\downarrow$  \\
\toprule
CenterPoint~\cite{yin2021center} &0.638   &0.555 &67.5\%  &760  \\
SimpleTrack~\cite{pang2023simpletrack} &0.668   &0.550 &70.3\%  &575 \\
\midrule
QD3DT~\cite{hu2022monocular} &0.217   &1.550 &37.5\%  &6856  \\
MUTR3D~\cite{zhang2022mutr3d} &0.270   &1.494 &41.1\%  &6018 \\
CC-3DT~\cite{fischer2022cc} &0.410  &1.274 &57.8\%   &3334 \\
PolarDETR~\cite{chen2022polar}  &0.273  &1.185 &40.4\%  &2170 \\
UVTR~\cite{li2022unifying} &0.519  &1.125 &59.9\%  &2204 \\
QTrack~\cite{yang2022quality}&0.480  &1.100 &59.7\%  &1484\\
Sparse4D~\cite{lin2022sparse4d} &0.519  &1.078 &63.3\%  &1090 \\
ByteTrackv2~\cite{zhang2023bytetrackv2} &0.564
  &1.005 &63.5\%  &704 \\
PF-Track~\cite{pang2023standing} &0.434 &1.252 &53.8\% &\textbf{249} \\ 
\rowcolor[gray]{.9} 
StreamPETR &\textbf{0.653 } &\textbf{0.876} &\textbf{73.3\%} &1037 \\
\bottomrule

\end{tabular}}
\vspace{-0.4cm}
\end{table}
\subsection{Implementation Details}
We conduct experiments with ResNet50~\cite{he2016resnet}, ResNet101, V2-99~\cite{lee2019energy} and ViT~\cite{dosovitskiy2020image} backbones under different pre-training. Following previous methods~\cite{li2022bevformer, liu2022petr, park2022time}, the performance of ResNet50 and ResNet101 models with pre-trained weights ImageNet~\cite{deng2009imagenet} and nuImages~\cite{caesar2020nuscenes} are provided on the nuScenes val set. To scale up our method, we also report results on the nuScenes test set with V2-99 initialized from DD3D~\cite{park2021dd3d} checkpoint 
and ViT-Large~\cite{dosovitskiy2020image}.
Following BEVFormerv2~\cite{yang2022bevformer}, the ViT-Large~\cite{dosovitskiy2020image} is pre-trained on  Objects365~\cite{shao2019objects365} and COCO~\cite{lin2014microsoft} dataset.

StreamPETR is trained by AdamW~\cite{loshchilov2017decoupled} optimizer with a batch size of 16. The base learning rate is set to 4e-4 and the cosine annealing policy is employed. Only key frames are used during both training and inference.  All experiments are conducted without CBGS~\cite{zhu2019class} strategy. Our implementation is mainly based on Focal-PETR~\cite{wang2022focal}, which introduces auxiliary 2D supervision. The models in the ablation study are trained for 24 epochs, while trained for 60 epochs when compared with others. In particular, we only train 24 epochs for ViT-L~\cite{dosovitskiy2020image} to prevent over-fitting. For image and BEV data augmentation, we adopt the same methods as PETR~\cite{huang2021bevdet,liu2022petr}. We randomly skip 1 frame during the training sequence for temporal data augmentation~\cite{li2022bevformer}. 

\begin{table}
\centering
\caption{Comparison on the Waymo \texttt{val} set. $\ast$ The saving interval  $\tau$ is set to 5 during testing. $\ddag$ The saving interval $\tau$ is set to 1. }
\vspace{0.1cm}

\label{tab:waymo_val}
\tiny
\resizebox{0.475\textwidth}{!}{
\setlength{\tabcolsep}{2.0pt}
\begin{tabular}{c|c|c|c|c} 
\toprule
\textbf{Methods} & \textbf{Backbone} &\textbf{mAPL}$\uparrow$ &\textbf{mAP}$\uparrow$ & \textbf{mAPH}$\uparrow$ \\
\toprule
BEVFormer++ \cite{ZhiqiLi2023BEVFormer}& ResNet101-DCN & 0.361  & 0.522 & 0.481  \\
MV-FCOS3D++ \cite{wang2022mv} & ResNet101-DCN  & 0.379 & 0.522 & 0.484 \\
PETR-DN \cite{liu2022petr}& ResNet101  & 0.358 &0.502 & 0.462 \\
PETRv2 \cite{liu2022petrv2}& ResNet101  & 0.366 &0.519 & 0.479 \\
\rowcolor[gray]{.9} 
StreamPETR${}^\ast$ & ResNet101  & \textbf{0.399}  & \textbf{0.553} & 0.517 \\
StreamPETR${}^\ddag$& ResNet101  & 0.395  & 0.551 & \textbf{0.518} \\
\bottomrule
\end{tabular}}
\vspace{-0.5cm}

\end{table}

\subsection{Main Results}
\noindent\textbf{NuScenes Dataset.} We compare the proposed StreamPETR with previous state-of-the-art vision-based 3D detectors on the nuScenes val and test set. As shown in Tab.~\ref{tab:main_val_set}, StreamPETR shows superior performance on mAP, NDS, mASE, and mAOE metrics when adopting ResNet101 backbone with nuImages pretraining. Compared with the single frame baseline Focal-PETR, StreamPETR has  considerable improvements of 11.4\% mAP and 13.1\% NDS. The mATE of StreamPETR is 10.9\% better than Focal-PETR, indicating that our object-centric temporal modeling is able to improve both the accuracy of localization. With image resolutions of 256$\times$704 and adopting ResNet50 backbone, StreamPETR exceeds the state-of-the-art method (SOLOFusion) by 0.5 \% mAP and 0.6 \% NDS.  When we reduce the number of queries and apply nuImages pre-training, our method has 2.3 \% and 1.6 \% advantages in mAP and NDS. At the same time, the inference speed of StreamPETR is 1.8$\times$ faster.

 When we compare the performance on the test set in Tab.~\ref{tab:main_test_set} and adopt a smaller V2-99 backbone, StreamPETR can surpass SOLOFusion with  ConvNext-Base backbone by 1.0\% mAP and 1.7\% NDS. 
Scaling up the backbone to ViT-Large~\cite{dosovitskiy2020image}, StreamPETR achieves 62.0\% of mAP, 67.6\% of NDS, and 25.8\% of mAOE. Note that it is the first online multi-view method that achieves comparable performance with CenterPoint.

For 3D multi-object tracking task, we simply extend the multi-object tracking of CenterPoint~\cite{yin2021center} to the multi-view 3D setting. Owing to the exceptional detection and velocity estimation performance, StreamPETR significantly outperforms ByteTrackv2~\cite{zhang2023bytetrackv2} with an impressive margin of +8.9\% AMOTA in Tab.~\ref{tab:tracking}. 
Furthermore, StreamPETR excels over CenterPoint~\cite{yin2021center} in AMOTA, and demonstrates superior benefits in RECALL. 

\begin{figure}[t]
\centering
\includegraphics[scale=0.15]{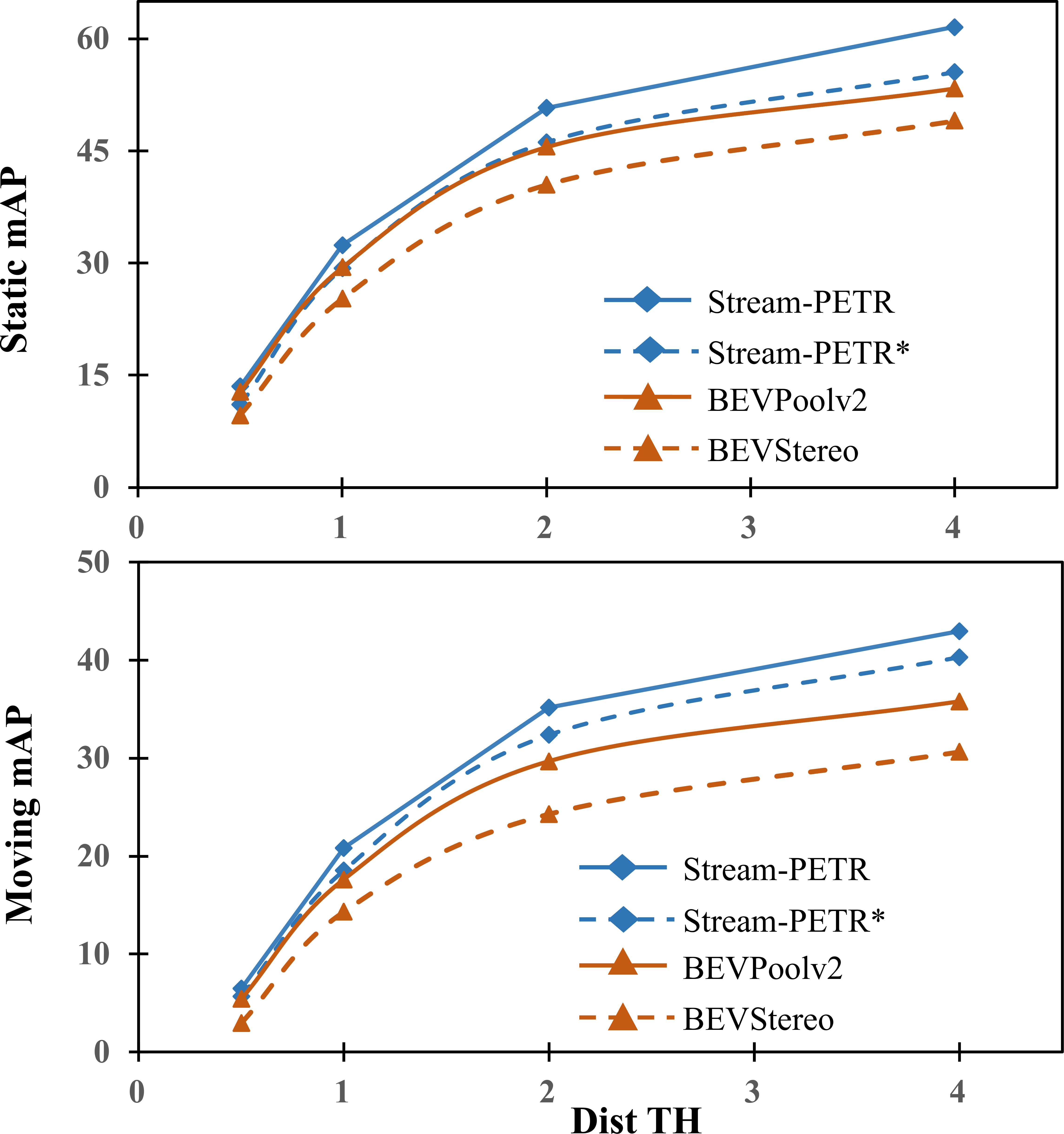}
\caption{The mAP results with different distance thresholds (Dist TH) on the nuScenes val set. $*$ indicates StreamPETR without the proposed motion-aware layer normalization. Top: Boxes with a velocity lower than 1m/s are maintained for analysis. Down: Boxes with a velocity higher than 1m/s are maintained for analysis.}
\label{moving_static_ap}
\vspace{-0.5cm}
\end{figure}

\noindent\textbf{Waymo Open Dataset.} 
In this section, we provide experimental results on the Waymo val set, as shown in Tab.~\ref{tab:waymo_val}. Our model has trained 24 epochs and the saving interval of the memory queue is set to 5. It can be seen that our method shows superiority in official metrics compared with the dense BEV methods, \eg BEVFormer++~\cite{ZhiqiLi2023BEVFormer} and MV-FCOS3D++~\cite{wang2022mv}. The Waymo open dataset has a larger evaluation range than nuScenes, our object-centric modeling method still shows obvious advantages in localization capability and longitudinal prediction. We also re-implemented PETR-DN and PETRv2 (all with query denoising~\cite{li2022dn}) as baseline models. StreamPETR outperforms the single-frame PETR-DN with a margin of 4.1\% mAPL, 5.1\% mAP, and 5.5\% mAP-H. The Waymo open dataset covers part of the horizontal FOV, while object-centric temporal modeling still brings significant improvement. When we adopt the checkpoint and adjust saving interval $\tau$ to 1 during testing, StreamPETR has slight performance degradation, proving the adaptability on sensor frequency.

\subsection{Ablation Study \& Analysis}
\noindent\textbf{Impact of Training Sequence Length.} 
StreamPETR is trained in local sliding windows and tested in online streaming video. To analyze the inconsistency between training and testing, we conduct experiments with varying numbers of training frames and show results in Tab.~\ref{tab:ablation_time_window}. When adding more training frames, the performance of StreamPETR continues to grow, and the performance gap between sliding windows and online video decreases obviously. It is worth noting that when the number of training frames increases to 8, video testing (40.2\% mAP, 50.5\% NDS) shows superior performance than the sliding window (39.6\% mAP, 50.1\% NDS), which proves that our method has a good potential to build long-term temporal dependency. Expanding to 12 frames brings limited performance improvement, so we train our models on 8 frames for experimental efficiency.

\noindent\textbf{Effect of Motion-aware Layer Normalization.}
We compare the different designs for decoupling the ego vehicle and moving objects in Tab.~\ref{tab:motion_ln_ablation}. It can be seen that the performance does not improve when adopting explicit motion compensation (MC). We argue that the explicit way may cause error propagation in the early training phase. The MLN implicitly encodes and decouples the movements of the ego vehicle and moving objects. Specifically, implicit encoding of ego poses has achieved significant improvements, among which mAP increases by 2.0\% and NDS increases by 1.8\%. Besides, the encoding of relative time offset $\triangle t$ and object velocity $v$ can further boost the performance. Both mAP and NDS are increased by 0.4\%, which indicates that dynamic properties have a beneficial effect on the temporal interaction between object queries.

\begin{table}
\centering
\caption{Training frames for long-term fusion. W indicates testing in the sliding window, and V indicates testing in online video.}
\label{tab:ablation_time_window}
\tiny
\resizebox{0.475\textwidth}{!}{
\setlength{\tabcolsep}{3pt}
\begin{tabular}{c|c|c|c|c|c}

\toprule
\textbf{Training frames} & 
\textbf{Test} & \textbf{mAP}$\uparrow$  &\textbf{NDS}$\uparrow$ & \textbf{mATE}$\downarrow$ & \textbf{mAVE}$\downarrow$  \\
\toprule
1  &-  &0.317  &0.372  &0.770  &0.885  \\
2  &\textbf {W}  &0.328  &0.410  &0.742  &0.726  \\
2  &\textbf {V} &0.315  &0.401  &0.738  &0.767 \\
4  &\textbf {W}    &0.377  &0.483  &0.683  &0.385  \\
4  &\textbf {V}  &0.366  &0.475  &0.685  &0.392  \\
8  &\textbf {W}    &0.396  &0.501  &0.664  &0.324  \\
\rowcolor[gray]{.9} 
8  &\textbf {V} &0.402  &0.505  &0.660  &\textbf{0.316}  \\
12 &\textbf {W}   &\textbf{0.403}  &0.507  &0.649  &0.325  \\
12 &\textbf {V}  &0.402  &\textbf{0.509}  &\textbf{0.645}  &\textbf{0.316}  \\
\bottomrule
\end{tabular}}
\vspace{-0.4cm}
\end{table}
\begin{table}
\centering
\caption{Ablation of motion-aware layer normalization. MC is explicit  motion compensation. LN is layer normalization.}
\label{tab:motion_ln_ablation}
\tiny
\resizebox{0.475\textwidth}{!}{
\setlength{\tabcolsep}{2.0pt}
\begin{tabular}{c|c|c|c|c|c|c|c|c} 
\toprule
\textbf{MC} &\textbf{LN} & \textbf{Ego Pose} & \textbf{Time} & \textbf{Velocity} & \textbf{mAP}$\uparrow$  &\textbf{NDS}$\uparrow$ & \textbf{mATE}$\downarrow$ & \textbf{mAVE}$\downarrow$  \\
\toprule
&  &  &  &  & 0.378 & 0.483 & 0.697 & 0.354  \\
\ding{52} &  &  &  &  & 0.380 & 0.481 & 0.693 & 0.379  \\
&\ding{52}  &  &  &  & 0.375 & 0.481 & 0.702 & 0.370  \\
&\ding{52}  &\ding{52}  & & & 0.398 & 0.501 & 0.667 & \textbf{0.316}  \\
&\ding{52}  &  &\ding{52} & & 0.381 & 0.488 & 0.697 & 0.354    \\
&\ding{52}  &  &  &\ding{52}  & 0.386 & 0.489 & 0.690 & 0.373  \\
\rowcolor[gray]{.9} 
&\ding{52}  &\ding{52}  &\ding{52}  &\ding{52}  & \textbf{0.402} & \textbf{0.505} & \textbf{0.660} & \textbf{0.316} \\
\bottomrule
\end{tabular}
}
\vspace{-0.4cm}
\end{table}
\begin{table}
\centering
\caption{Number of frames (N) for long-term fusion.}
\label{tab:ablation_memory_len}
\tiny
\resizebox{0.475\textwidth}{!}{
\setlength{\tabcolsep}{3.0pt}
\begin{tabular}{c|c|c|c|c|c}

\toprule
\textbf{number frames}& \textbf{mAP}$\uparrow$  &\textbf{NDS}$\uparrow$ & \textbf{mATE}$\downarrow$ & \textbf{mAVE}$\downarrow$ &\textbf{FPS}$\uparrow$ \\
\toprule
0  &0.317  &0.372  &0.770  &0.885& \textbf{27.7}   \\
1  & 0.394 & 0.501 & 0.669 &  0.324& \textbf{27.7} \\
2  &0.401  &\textbf{0.505}  & \textbf{0.660}  &\textbf{0.314}& 27.4   \\
3  & 0.400 & 0.504 & 0.663 & 0.322 & 27.3 \\
\rowcolor[gray]{.9} 
4  & \textbf{0.402} & \textbf{0.505} & \textbf{0.660} & 0.316& 27.1 \\
\bottomrule
\end{tabular}}
\vspace{-0.5cm}
\end{table}

\begin{figure*}[t]
\centering
\includegraphics[scale=0.1]{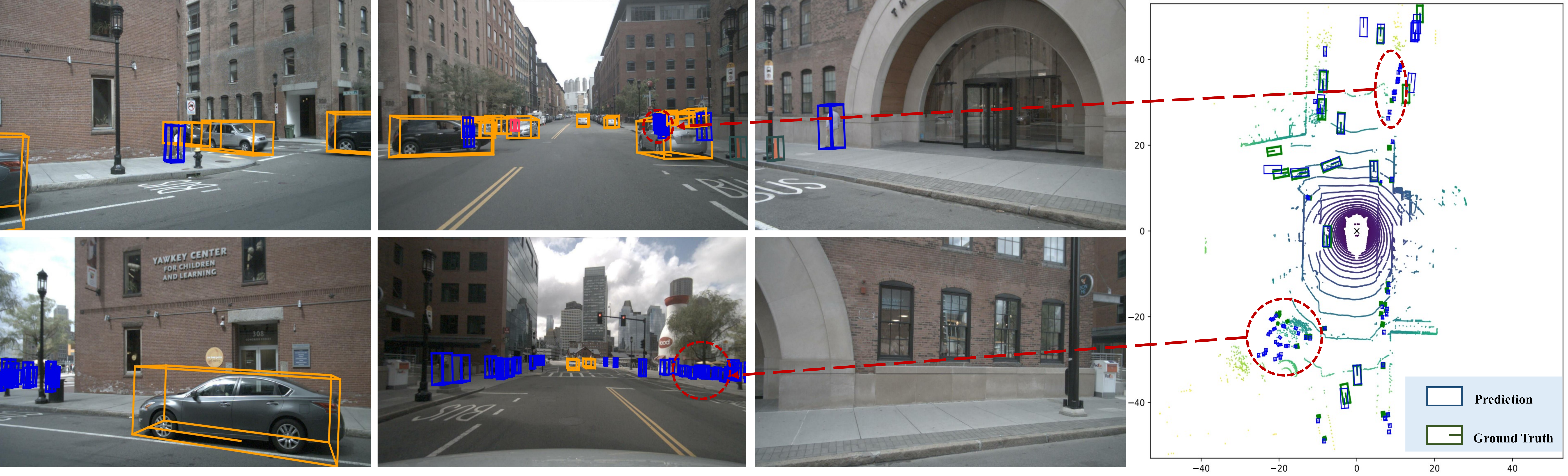}
\caption{Visualization results of StreamPETR. On the BEV plane (right), the groud-truth and predictions are drawn in green and blue rectangles respectively. The failure cases are marked by red circles.}
\label{scene_vis}
\vspace{-0.4cm}
\end{figure*}

\noindent\textbf{Number of Frames for Long-term Fusion.}
In Tab.~\ref{tab:ablation_memory_len}, we analyze the impacts of memory size on hybrid attention. We can find that the mAP and NDS are improved with the increase of the memory size and begin to saturate when reaching 2 frames (nearly 1 second). The object query in StreamPETR is propagated and updated recursively, so even without a large-capacity memory queue, our method can still build a long-term spatial-temporal dependency. Since increasing the memory queue brings negligible computing costs, we use 4 frames to alleviate forgetting and obtain more stable results.

\noindent\textbf{Perspective v.s. Object-Centric.}
StreamPETR achieves efficient temporal modeling through the interaction of sparse object queries. An alternative solution is to build temporal interaction via the perspective memory~\cite{liu2022petrv2}. As shown in Tab.~\ref{tab:ablation_dual_memory}, the query-based temporal modeling has superior performance than perspective-based both on speed and accuracy. The combination of the query and perspective memory does not further improve the performance, implying that the temporal propagation of global query interaction is sufficient to achieve leading performance. Besides, concatenating current object queries with the queries of the last frame improves 0.7\% mAP and 0.9\% NDS.

\noindent\textbf{Analysis of Moving Objects.} In this section, we detailed analyze the performance of StreamPETR on perceiving static and moving objects respectively. For fair comparisons, all models are trained with 24 epochs without CBGS~\cite{zhu2019class} and evaluated on the nuScenes ~\cite{caesar2020nuscenes} val set. The detection performance of moving objects still lags behind that of static objects to a large margin even with temporal modeling. Compared with dense BEV paradigms~\cite{li2022bevstereo, huang2022bevpoolv2}, StreamPETR$*$ has reached promising performance on both static and moving objects. This proves the superiority of object-centric temporal modeling, which has global temporal and spatial receptive fields. Applying the implicit encoding for motion information, the performance of StreamPETR can be further improved.

\begin{table}
\centering
\caption{From of the temporal propagation. 'Perspective and Object' mean propagating temporal information via image features and object queries respectively. 'Propagated' indicates concatenating the propagated queries from last frame.}
\vspace{2pt}
\label{tab:ablation_dual_memory}
\tiny
\resizebox{0.475\textwidth}{!}{
\setlength{\tabcolsep}{2.0pt}
\begin{tabular}{c|c|c|c|c|c|c|c}

\toprule
\textbf{Perspective} & \textbf{Object}&  \textbf{Propagated} & \textbf{mAP}$\uparrow$  &\textbf{NDS}$\uparrow$ & \textbf{mATE}$\downarrow$ & \textbf{mAVE}$\downarrow$  & \textbf{FPS}$\uparrow$ \\
\toprule
  &   &&0.317  &0.372  &0.770  &0.885 & \textbf{27.7}\\
   \ding{52} & &\ding{52} & 0.361 &0.459&0.731& 0.374 & 18.9\\
& \ding{52} &  &0.395  &0.496  &0.703  &0.363 & 27.1\\
\rowcolor[gray]{.9} 
  & \ding{52} &  \ding{52} & \textbf{0.402}  & \textbf{0.505}  & \textbf{0.660}  & \textbf{0.316} & 27.1\\
\ding{52}  & \ding{52}  &\ding{52}& \textbf{0.402} & 0.503 & 0.662 & 0.341 & 18.6\\
\bottomrule
\end{tabular}
}
\vspace{-0.5cm}
\end{table}
\subsection{Failure Cases}
We show the detection results of a challenging scene in Fig.~\ref{scene_vis}. 
StreamPETR shows impressive results on crowded objects within the detection range of 30m. However, our method has many False Positives on remote objects. It is a common phenomenon of camera-based methods.
In a complex urban scene, the duplicated predictions on remote objects can be tolerable and cause relatively little impact.

\section{Conclusion}
In this paper, we propose StreamPETR, an effective long-sequence 3D object detector. Different from the previous works, our method explores an object-centric paradigm that propagates temporal information through object queries frame by frame. In addition, a motion-aware layer normalization is adopted to introduce the motion information. StreamPETR achieves leading performance improvements while introducing negligible storage and computation cost. It is the first online multi-view method that achieves comparable performance with lidar-based methods. We hope StreamPETR can provide some new insights into long-sequence modeling for the community.


\section*{A. Appendix}
\subsection*{A.1. Algorithm Workflow}

\begin{algorithm}
    \small
    \caption{Propagation Transformer}
    \hspace*{\algorithmicindent} 
    \textbf{Input}: Multi-view 2D features from streaming video $\mathbi F_{2d} =\{F_{2d}^1, F_{2d}^2, \cdots, F_{2d}^T\}$. A set of learnable reference points $Q_{p}^{init}$.  \\
    \hspace*{\algorithmicindent} \textbf{Memory Queue}: Memory queue of $N$ historical frames $\mathbi{X} = \{X^{t-N}, \cdots, X^{t-2}, X^{t-1}\}$. For each time stamp ${t-k}$, we maintain the query states with motion information, $X^{t-k} = \{\triangle t^{t-k}, Q_{c}^{t-k}, Q_{p}^{t-k}, v^{t-k}, E^{t-k}\}$.\\
    \hspace*{\algorithmicindent} \textbf{Output}: 3D bounding boxes prediction with classification scores $\mathbi{b}^{t} = (x, y, z, l, w, h, \theta, v_x, v_y, cls)$.
    \begin{algorithmic}[1]
    \State $\mathbi{X} \leftarrow \O$
    \For{$t \in \mathbi{T}$}
    
        \!\!\!\!\! \textbf{(1) Motion compensation}
        \State $ {[\tilde{Q}_{p}^{t-N:t-1}, \tilde{E}_{t-N:t-1}^{t}]} \leftarrow Ego([Q_{p}^{t-N:t-1}, E^{t-N:t-1}])$ 
        \State $\tilde{Q}_{pe}^{t-N:t-1} \leftarrow \psi(\tilde{Q}_{p}^{t-N:t-1})$ 
        \State $M \leftarrow [\tilde{E}_{t-N:t-1}^{t}, \triangle t^{t-N:t-1}, v^{t-N:t-1}]$
        \State $\tilde{Q}_{pe}^{t-N:t-1} \leftarrow MLN(\tilde{Q}_{pe}^{t-N:t-1}, M)$
        \State $\tilde{Q}_{c}^{t-N:t-1} \leftarrow MLN(Q_{c}^{t-N:t-1}, M)$

        \!\!\!\!\! \textbf{(2) Propagate query}
        \State $Q_{pe}^{init} \leftarrow \psi(Q_{p}^{init})$ 
        \State $Q_{c}^{init} \leftarrow Zero\_Like(Q_{pe}^{init})$ 
        \State $Q_{pe}^{0}\;\;\leftarrow Concat([Q_{pe}^{init}, \tilde{Q}_{pe}^{t-1}])$
        \State $Q_{c}^{0}\;\:\enspace\leftarrow Concat[Q_{c}^{init}, \tilde{Q}_{c}^{t-1}]$

        \!\!\!\!\! \textbf{(3) Spatial-temporal interaction}
        \For{$i \in L$}
            \State $\tilde{Q}_{pe}^{hybrid} \leftarrow Concat([Q_{pe}^{i}, \tilde{Q}_{pe}^{t-N:t-1}])$
            \State $\tilde{Q}_{c}^{hybrid} \leftarrow Concat[Q_{c}^{i}, \tilde{Q}_{c}^{t-N:t-1}]$
            \State $Q_{c}^{i}\quad\leftarrow Hybrid\_Attn([Q_{c}^{i}, Q_{pe}, \tilde{Q}_{c}^{hybrid}, \tilde{Q}_{pe}^{hybrid}]) $
            \State $Q_{c}^{i+1} \leftarrow Cross\_Attn([Q_{c}^{i},Q_{pe}, F_{2d}^t, F_{3d\_pe}^t])$
            \State $Q_{pe}^{i+1} \leftarrow Q_{pe}^{i}$ 
        \EndFor

        \!\!\!\!\! \textbf{(4) Update memory queue}
        \State$\mathbi{b}^{t} \leftarrow Head(Q_{c}^{L})$
        \State$ index \leftarrow TopK(\mathbi{b}^{t})$
        \State ${X}^{t} \leftarrow Gather(index)$ 
        \State $\mathbi{X} \,\,\leftarrow \{X^{t-N+1}, \cdots, X^{t-1}, X^{t}\}$ 
    \EndFor
    \end{algorithmic}
    \label{alg:streampetr}
\end{algorithm}
The workflow of our proposed Propagation Transformer is shown in Alg.~\ref{alg:streampetr}, which is divided into four stages:

(1) The motion compensation takes the information of the memory queue as input (including the relative time interval $\triangle t$, context embedding $Q_{c}$, object center $Q_{p}$, velocity $v$, and ego-pose matrix $E$). The object 3D centers ${Q}_{p}^{t-N:t-1}$ in the memory queue are explicitly aligned to the current frame according to the ego pose $Ego$. Then the aligned centers $\Tilde{Q}_{p}^{t-N:t-1}$ are used to generate position encoding of object query by a single MLP layer $\psi$. Afterward, we apply the Motion aware Layer Normalization (MLN) to encode motion information $M$ into the historical queries.

(2) The generation of the object queries is mainly based on the learnable query embedding $[Q_{pe}^{init}, Q_{c}^{init}]$ and the obtained propagated query embedding $[\Tilde{Q}_{pe}^{t-1}, \Tilde{Q}_{c}^{t-1}]$ from the last frame $t-1$.

(3) The spatial-temporal interaction of the Propagation Transformer has stacked L layers of the hybrid attention ($Hybrid\_Attn$) and cross attention ($Cross\_Attn$). For each layer, the hybrid attention performs the interaction of current queries $[Q_{c}^{i}, Q_{pe}^{i}]$ and historical queries $[\Tilde{Q}_{c}^{hybrid}, \Tilde{Q}_{pe}^{hybrid}]$, and the cross attention performs the interaction of current queries and image tokens $[F_{2d}^t, F_{3d\_pe}^t]$. $F_{3d\_pe}^t$ is the 3D position encoding proposed in PETR~\cite{liu2022petr}.

(4) After the layer-by-layer refinement, a 3D Head ($Head$) are conducted to generate the predictions. Then we select top-K foreground objects according to the classification scores and push the information of the selected objects to the memory queue \mathbi{X}. 
\begin{table}
\centering
\caption{Flash Attention for efficient training (V2-99~\cite{lee2019energy} backbone with input resolution of $1600\times640$). }
\label{tab:fash_attn}
\tiny
\resizebox{0.475\textwidth}{!}{
\begin{tabular}{c|c|c} 
\toprule
\textbf{Flash Attn} & \textbf{A100 Training Time (s/iter)} $\downarrow$& \textbf{GPU Memory (G)}$\downarrow$  \\
\midrule
\ding{52} & \textbf{1.51} & \textbf{27G} \\
\ding{56} & 1.68  & 61G   \\
\bottomrule
\end{tabular}}
\end{table}
\subsection*{A.2. Additional Details}
Qualitative results of StreamPETR are provided in \href{https://drive.google.com/file/d/1vNY2o6-QwR1dWj8H7BdzTR0VDSXMoJcY/view?usp=sharing}{video}.

Here we provide more details for reproducing the results. First, we detach the gradient of the first 6 frames and compute the gradient and losses of the last 2 frames,  which can accelerate the convergence. We additionally adopt Flash Attention~\cite{dao2022flashattention} to further save the GPU memory, as shown in Tab.~\ref{tab:fash_attn}. The query denoising~\cite{li2022dn} is also conducted following PETRv2~\cite{liu2022petrv2}. When measuring the inference speed, we close the Flash Attention.

\begin{table}
\centering
\caption{Applicability of our method. We extend object-centric temporal modeling to DETR3D.}
\label{tab:ablation_detr3d}
\tiny
\resizebox{0.475\textwidth}{!}{
\setlength{\tabcolsep}{2.5pt}
\begin{tabular}{c|c|c|c|c|c} 
\toprule
\textbf{Method} &\textbf{mAP}$\uparrow$ &\textbf{NDS}$\uparrow$ & \textbf{mATE}$\downarrow$ &\textbf{mAVE}$\downarrow$  & \textbf{FPS}$\uparrow$ \\
\toprule
DETR3D & 0.347  & 0.422 & 0.765 & 0.876 & \textbf{6.3}  \\
\rowcolor[gray]{.9} 
    Stream-DETR3D & \textbf{0.396}  & \textbf{0.490} & \textbf{0.723} & \textbf{0.487}  & 6.2  \\
\bottomrule
\end{tabular}}
\vspace{-0.5cm}
\end{table}
\subsection*{A.3. Extension of Our Method}
To verify the extensibility of our method, we conduct experiments on another sparse query base model DETR3D~\cite{wang2022detr3d}. We use ResNet101-DCN as the backbone, without additional augmentation and CBGS~\cite{zhu2019class}. Results in Tab.~\ref{tab:ablation_detr3d} show that StreamDETR3D brings 4.9\%  and 6.8\% improvements on mAP and NDS, while the inference speed is little impacted. Compared with the PETR paradigm, the improvement of DETR3D is relatively small. One possible reason is that the local spatial attention
adopted by DETR3D limits the performance.

{\small
\bibliographystyle{ieee_fullname}
\bibliography{egbib}
}
\end{document}